\documentclass[letterpaper, 10 pt, conference]{IEEEtran}
\IEEEoverridecommandlockouts
\usepackage{balance}
\usepackage{cite}
\usepackage{amsmath,amssymb,amsfonts,bbm,mathtools}
\usepackage{hyperref}
\usepackage[ruled]{algorithm2e}
\usepackage{graphicx,subcaption}
\usepackage{textcomp}
\usepackage{xcolor}
\usepackage{comment}
\usepackage{multirow}
\def\BibTeX{{\rm B\kern-.05em{\sc i\kern-.025em b}\kern-.08em
    T\kern-.1667em\lower.7ex\hbox{E}\kern-.125emX}}
\begin{document}

\title{\LARGE \bf Markerless Suture Needle 6D Pose Tracking with Robust Uncertainty Estimation for Autonomous Minimally Invasive Robotic Surgery}

\author{
Zih-Yun Chiu$^1$, Albert Z Liao$^1$, Florian Richter$^1$ \IEEEmembership{Student Member, IEEE}\\Bjorn Johnson$^2$, and Michael C. Yip$^1$ \IEEEmembership{Senior Member, IEEE}
\thanks{This research was supported by the Telemedicine and Advanced Technology Research Center, an NSF GRFP, and NSF \#2045803.}
\thanks{$^1$Zih-Yun Chiu, Albert Z Liao, Florian Richter, and Michael C. Yip are with the Department of Electrical and Computer Engineering, University of California San Diego, La Jolla, CA 92093 USA. {\tt\small \{zchiu, azliao, frichter, yip\}@ucsd.edu}}%
\thanks{$^2$Bjorn Johnson is with the Department of Computer Science and Engineering, University of California San Diego, La Jolla, CA 92093 USA. {\tt\small \{bljohnso\}@ucsd.edu}.}%
}

\maketitle
\begin{abstract}
Suture needle localization is necessary for autonomous suturing.
Previous approaches in autonomous suturing often relied on fiducial markers rather than markerless detection schemes for localizing a suture needle due to the inconsistency of markerless detections.
However, fiducial markers are not practical for real-world applications and can often be occluded from environmental factors in surgery (e.g., blood).
Therefore in this work, we present a robust tracking approach for estimating the 6D pose of a suture needle when using inconsistent detections.
We define observation models based on suture needles' geometry that captures the uncertainty of the detections and fuse them temporally in a probabilistic fashion.
In our experiments, we compare different permutations of the observation models in the suture needle localization task to show their effectiveness. 
Our proposed method outperforms previous approaches in localizing a suture needle.
We also demonstrate the proposed tracking method in an autonomous suture needle regrasping task and ex vivo environments.
\end{abstract}

\section{Introduction}
In recent years, there has been a growing interest in achieving autonomous suturing on robotic surgical systems because it is a time-consuming task and particularly tedious and challenging in Minimally Invasive Surgeries~\cite{garcia1998manual, hubens2003performance}.
The complexity of the task directly leads to the development of a variety of necessary components to automate suturing,
which include identifying entry and exit points for a needle~\cite{liu2016needle}, suture path planning~\cite{jackson2013needle}, needle regrasping~\cite{chiu2020bimanual}, and knot tying~\cite{chow2013improved}.
The crucial component that we will be focusing on in this work is perception for automating suturing.

The perception necessary to attain autonomous suturing can be broken into two categories: environment reconstruction and surgical tool localization.
The environment reconstruction refers to techniques that can provide 3D information about the tissue to be sutured \cite{li2020super, lu2020super, long2021dssr}.
The surgical tool localization in tandem would provide a complete geometric description of the surgical tools, i.e., needle drivers, suture needles, thread, in the same frame of reference as the environment reconstruction.
Surgical robotic tool tracking methods leverage the kinematic and joint encoder readings for high accuracy localization in the camera frame \cite{zhao2015efficient, original_rcs, richter2021robotic}.
Meanwhile, suture thread and needles are manipulated by surgical tools hence a reliance on endoscopic camera data is required to localize it.

\begin{figure}[t]
    \centering
    \vspace{2mm}
    \includegraphics[width=\linewidth]{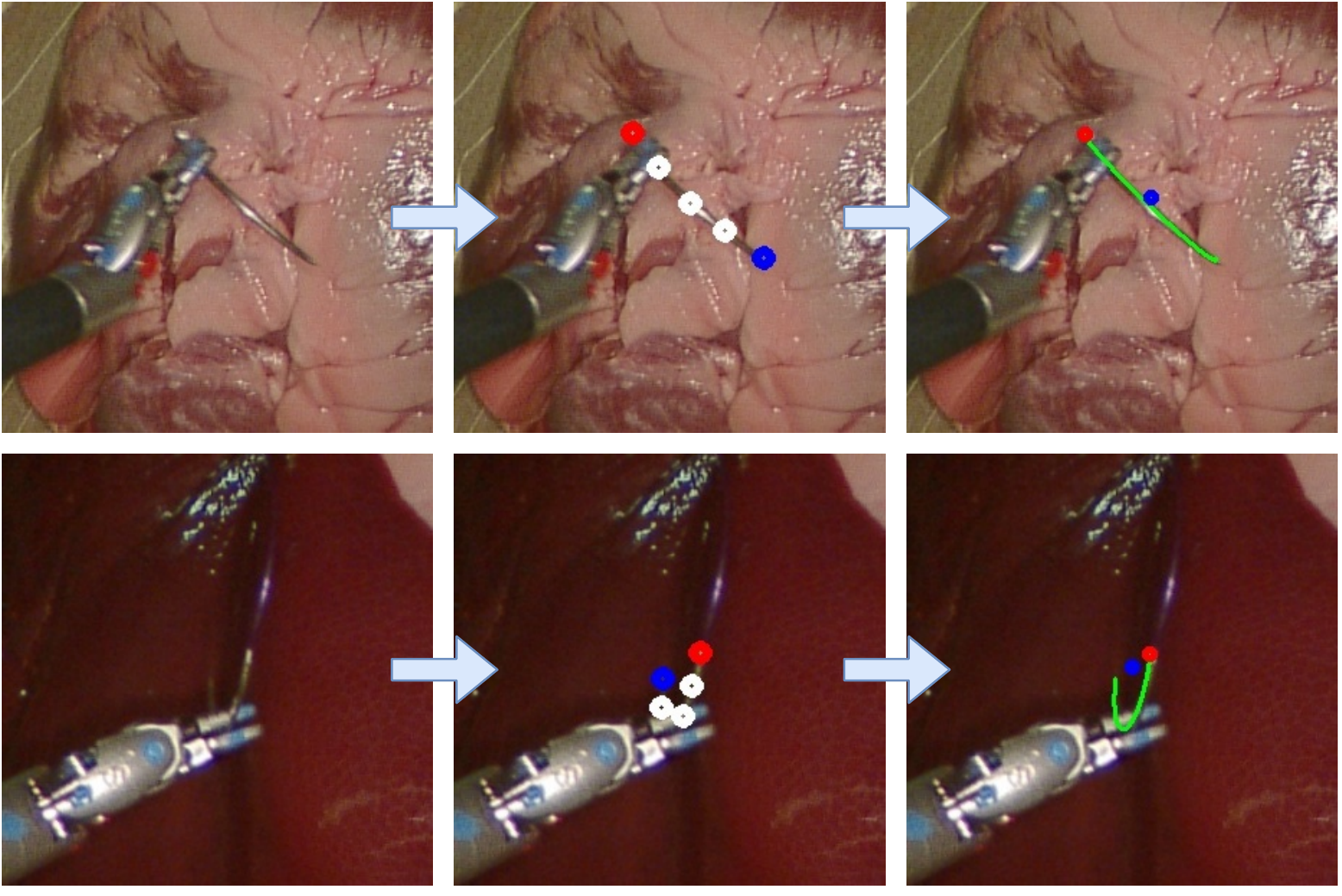}
    \caption{Raw image (left column), markerless feature detection (middle column), and projection of the tracked needle pose (right column). 
    In the right column images, the green curve is the tracked needle body, the red dot is the tracked endpoint of the needle connected to the suture thread, and the blue dot is the tracked center of the needle's circle.
    The tracked projection overlaps well with the real one, indicating our method's effectiveness in ex vivo environments.}
    \label{fig:cover_image}
\end{figure}

\subsection{Related Work}

With an endoscopic camera detecting the needle, how the images are used to reconstruct the 6D pose of a suture needle is the critical component for robust needle localization.
In~\cite{iyer2013single}, the needle pose is reconstructed from an image at each time step using the method proposed in~\cite{lo2002trip};
\cite{d2018automated} reconstructed the needle pose from the filtered detection of the markers;
\cite{wilcox2021learning} segmented out the needle by training a fully convolutional neural network and reconstructed the pose by fitting a circle to a point cloud with RANSAC. 
However, the approaches do not capture the motions and may fail under large needle movements. 

Previous work also applies Bayesian filters to localize a suture needle due to their ability to model uncertainty in the images and the motions. 
The main differences between these approaches are their choices of the observations extracted from the images. 
In~\cite{ferro2017vision}, the observation is the needle pose reconstructed with the same method used in~\cite{iyer2013single}, but Extended Kalman Filter (EKF) is applied to track the pose over time. 
A similar idea can be found in~\cite{sen2016automating}, where the observation is the needle pose reconstructed from an ellipse fitted with the point set registered between real and rendered images.
Fitting an ellipse to a noisy needle segmentation is usually inaccurate, so these approaches rely on markers to obtain a relatively complete needle segmentation.
Nevertheless, in real-world surgeries where markerless needle detections are preferred, these observation models are more likely to get inaccurate results due to noisy segmentation. 
Therefore, we propose the observation models that do not involve the reconstructed detections from the original detected points since they make it more challenging to model the uncertainty. 
Instead, our observation models use only the original point detections and modeling the uncertainty directly from these detections.

Besides observing the reconstructed needle pose, the similarity between the actual and rendered images is also considered an observation in previous work. 
In~\cite{kurose2013preliminary}, a database of numerous poses of a needle and their corresponding 2D projections are rendered beforehand. 
Then during run time, the likelihood between the current needle segmentation and the projections in the database are tracked by Particle Filter (PF). 
Instead of collecting a database in advance, \cite{ozguner2018three} calculated a matching score between the current and each rendered needle segmentation in real-time, and this score is used to update the weight of each particle in PF.
Calculating a reliable matching score between noisy and perfectly rendered needle projections requires some form of registration between two sets of points~\cite{sen2016automating, kurose2013preliminary, ozguner2018three}. 
The reliability increases with more needle points rendered, but increasing the number of projected points reduces the overall computational speed~\cite{ozguner2018three}.
To prevent the necessity of point-set registration, one of our proposed observation models, \textit{Points Matching to Ellipse Observation}, measures how far the detected points are from the estimated needle body. 
This measurement does not require any registration process and can work with only a few point detections. 

\subsection{Contributions}

In this work, we present a robust tracking approach for estimating the 6D pose of a suture needle with inconsistent detections, which occurs when using markerless detection schemes.
Our tracking method overcomes this challenge by modeling the uncertainty of the detections and fusing the observations temporally in a probabilistic fashion with PF.
The observation models of the detections are derived from suture needles' geometry. 
We present a novel observation model named \textit{Points Matching to Ellipse Observation}, which generalizes to any detected pixel of a suture needle and does not require an associated 3D coordinate on the needle.
Furthermore, we incorporate a motion model to estimate the suture needle's motion between image frames by incorporating the motion of the surgical tool that is grasping the suture needle.

Our experimental results in simulation show that the novel \textit{Points Matching to Ellipse} observation model achieves robust tracking and outperforms previous work in suture needle localization.
Moreover, we demonstrate that tracking with this observation model is robust enough to achieve a high success rate in our previously developed needle regrasping policy~\cite{chiu2020bimanual}. 
The integration of a markerless feature detector and the proposed tracking method is also tested in real-world environments. 
Several examples of detected features and the projection of the tracked pose are shown in Fig. \ref{fig:cover_image}. 
The consistency between the tracked projection and the real one indicates the effectiveness of our method under different scenarios, including occlusion and ex-vivo environments.

\begin{algorithm}[t!]
\footnotesize
  \SetAlgoLined
  \KwIn{initial pose $\mathbf{p}_0$ and noise covariance $\Sigma_0$, robot actions $\mathbf{a}_{1:T}$, motion noise covariance $\Sigma_m$, image observations $\mathbb{I}_{1:T}$, observation noise covariance $\Sigma_o$}
  \KwOut{tracked needle pose $\mathbf{p}_{1:T}$}
  \tcp{Initialize Distribution}
  $\{ \alpha^{(i)}, \mathbf{p}^{(i)}_{0|0} \}_{i=1}^{N_s} \leftarrow sampleInitial(\mathbf{p}_0, \Sigma_0)$\\
  \For{timestep $t = 1,\dots, T$}{
    \tcp{Detect Needle Features}
      $\mathbf{o}_t \leftarrow getDetections(\mathbb{I}_t)$\\
    \For{particle $i = 1,\dots,N_s$}{
        \tcp{Predict and Update}
    $\mathbf{p}^{(i)}_{t|t-1} \leftarrow 
        sampleMotion(\mathbf{p}^{(i)}_{t-1|t-1}, \mathbf{a}_t, \Sigma_m)$\\
    $\alpha^{(i)} \leftarrow 
        \alpha^{(i)} \cdot probObsModel(\mathbf{p}^{(i)}_{t|t-1}, \mathbf{o}_t, \Sigma_o)$
        }
    \tcp{Normalize Particle Distribution}
    $\mathbf{p}^{(i)}_{t|t} \leftarrow \mathbf{p}^{(i)}_{t|t-1}$\\
    $\{ \alpha^{(i)} \}_{i=1}^{N_s} \leftarrow normalizeWeights \left( \{ \alpha^{(i)} \}_{i=1}^{N_s} \right)$\\
        \tcp{Stratify Resampling \cite{kitagawa1996monte}}
    \If{$effectiveParticles(\{ \mathbf{p}^{(i)}_{t|t} \}_{i=1}^{N_s}) < N_{eff}$}{
        $\{ \mathbf{p}^{(i)}_{t|t} \}_{i=1}^{N_s} \leftarrow resampling(\{ \mathbf{p}^{(i)}_{t|t} \}_{i=1}^{N_s})$\\
    }
    \tcp{Return Mean as Needle Pose}
    $\mathbf{p}_{t} \leftarrow \sum \limits_{i=1}^{N_s} \alpha^{(i)} \mathbf{p}^{(i)}_{t|t} $\\
  }
  \caption{Needle Tracking with PF}
  \label{alg:particle_filter}
\end{algorithm}

\section{Methods}

Our goal is to track the pose of the suture needle in the camera frame, $\mathbf{p}_{t} = \left[ \mathbf{b}_t^\top\ \mathbf{q}_t^\top \right]^\top$, at each time step $t$, where $\mathbf{b}_{t} \in \mathbb{R}^3$ is the position, and $\mathbf{q}_{t} \in \mathbb{R}^3$ is the axis-angle orientation. 
Bayesian state estimation methods, i.e., Bayesian filters, are used to track the needle pose since they consider the uncertainty in the motions and observations.
A motion model, detection technique, and observation model need to be defined to track an object using Bayesian filters.
Our approaches to these components are detailed in the coming sections.
An outline of our Particle Filter (PF) implementation to solve the Bayesian state estimation of the suture needle pose using these components is shown in  Algorithm \ref{alg:particle_filter}.


\subsection{Motion Model}

An action that predicts the suture needle's motion can be predefined, read from the sensors of a robot, or derived by tracking the surgical needle driver holding the needle~\cite{richter2021robotic}.
Let the suture needle's action be denoted as $\mathbf{a}_t = [ \mathbf{a}_{b,t} \  \mathbf{a}_{q,t} ]^\top$ at time $t$ where $\mathbf{a}_{b,t} \in \mathbb{R}^3$ represents the action of position, and $\mathbf{a}_{q,t} \in \mathbb{R}^3$ represents the action of orientation. 
The motion model used for prediction in Algorithm \ref{alg:particle_filter} is then written as
\begin{equation}
    \begin{bmatrix}
        \mathbf{b}_{t} \\
        \mathbf{q}_{t}
    \end{bmatrix} = 
    \begin{bmatrix}
        \mathbf{b}_{t-1} + \mathbf{a}_{b,t} \\
        \mathbf{a}_{q,t} \circ \mathbf{q}_{t-1}
    \end{bmatrix} + \mathbf{w}_t, 
\end{equation}
where $\mathbf{a}_{q,t} \circ \mathbf{q}_{t-1}$ is the composition of two axis-angle orientations~\cite{altmann1989hamilton}, and $\mathbf{w}_t \sim \mathcal{N}\left( \mathbf{0}, \Sigma_m \right)$, $\Sigma_m \in \mathbb{R}^{6 \times 6}$, is the motion noise.
Gaussian noise is used for the motion model noise due to its ability to generalize over a wide range of distributions.

\subsection{Feature Extraction of Suture Needle}
\label{subsec:ob_feature}

\begin{figure}[t]
    \centering
    \vspace{2mm}
    \begin{subfigure}{0.46\linewidth}
        \centering
        \includegraphics[width=\textwidth]{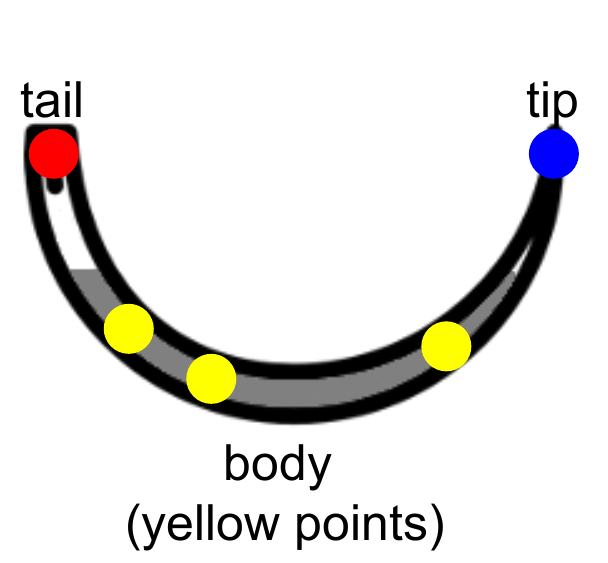}
        \caption{Point feature}
        \label{fig:point_feature_example}
    \end{subfigure}
    \begin{subfigure}{0.46\linewidth}
        \centering
        \includegraphics[width=\textwidth]{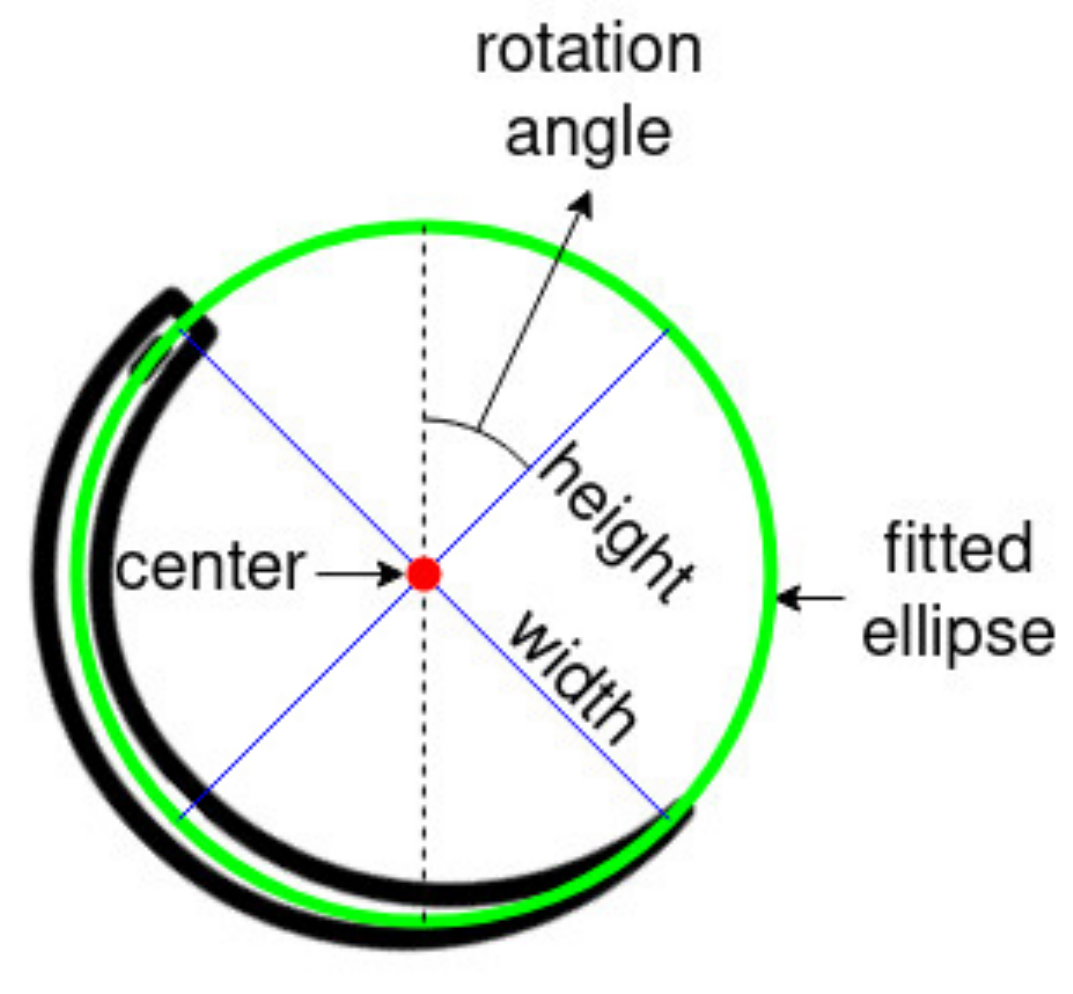}
        \caption{Ellipse parameter feature}
        \label{fig:ellipse_feature_example}
    \end{subfigure}
    \caption{An example of point and ellipse parameter features. Note that in the left figure, the body points can be randomly distributed on the gray area.}
    \vspace{-0.5mm}
    \label{fig:feature_example}
\end{figure}

Suture needles are detected from endoscopic images, and this can be done in a variety of ways, including in a markerless fashion (e.g., \cite{iyer2013single, d2018automated, wilcox2021learning, ferro2017vision, sen2016automating, kurose2013preliminary, ozguner2018three, speidel2015image, mathis2018deeplabcut}).
Ultimately, the detections are a list of pixel coordinates denoted as needle points: $\mathbf{f}_{t,i} \in \mathbb{R}^2$ for $i=1,\dots,N_f$ at time step $t$ where there are a total of $N_f$ points. 
We present two features derived from the needle points:

\subsubsection{Point feature}
The detected needle points can be directly used as an observation. 
Therefore, the point feature can be simply defined as $\mathbf{o}_{p,t,i} = \mathbf{f}_{t,i}$. 
An example of a point observation is shown in Fig. \ref{fig:point_feature_example}. 

\subsubsection{Ellipse parameter feature}
Since the projection of a circular suture needle onto the image plane forms a part of an ellipse, the parameters of this ellipse can be used as another observation. 
Let the observation at time $t$ be $\mathbf{o}_{ep,t} \in \mathbb{R}^5$, which includes the center pixel coordinate, width and height measured in pixels, and the rotation of the ellipse along the axis perpendicular to the image plane.
To get $\mathbf{o}_{ep,t}$, we first consider fitting the detected needle points with the ellipse's general equation: 
\begin{equation}
    \label{eq:general_ellipse_eq}
    a_t x_t^2 + 2b_t x_t y_t + c_t y_t^2 + 2d_t x_t + 2e_t y_t + 1 = 0, 
\end{equation}
where $\left[ a_t\ b_t\ c_t\ d_t\ e_t \right]^\top \in \mathbb{R}^5$ are the coefficients of the ellipse, and $\left[ x_t\ y_t \right]^\top \in \mathbb{R}^2$ is the pixel coordinate of a point on the ellipse. 
The detected needle points, $\mathbf{f}_{t,i} = \left[ \mathbf{x}_{t,i}\ \mathbf{y}_{t,i} \right]^\top$, are used to solve for the ellipse coefficients.
The ellipse coefficients can be solved for using the following linear equation:
\begin{equation}
    \mathbf{D}_t
    \begin{bmatrix}
        a & b & c & d & e
    \end{bmatrix}^\top = 
    \begin{bmatrix}
        -1 & -1 & -1 & -1 & -1
    \end{bmatrix}^\top,  
\end{equation}
where
\begin{equation}
    \mathbf{D}_t = 
    \begin{bmatrix}
        x_{t,1}^2 & 2 x_{t,1} y_{t,1} & y_{t,1}^2 & 2 x_{t,1} & 2 y_{t,1} \\
        & & \vdots & & \\
        x_{t,N_f}^2 & 2 x_{t,N_f} y_{t,N_f} & y_{t,N_f}^2 & 2 x_{t,N_f} & 2 y_{t,N_f}
    \end{bmatrix}.
\end{equation}
By the standard method of solving a system of linear equations in the matrix form, 
\begin{equation}
    \begin{bmatrix}
        a & b & c & d & e
    \end{bmatrix}^\top = 
    \mathbf{D}_t^{-1}
    \begin{bmatrix}
        -1 & -1 & -1 & -1 & -1
    \end{bmatrix}^\top,
    \label{equ:ellipse_coeff_fitting}
\end{equation}
where an exact inverse can exist if $N_f = 5$, i.e. 5 detected points.
Otherwise, a pseudo-inverse can be used, and the best fit will be solved.
From the fitted ellipse coefficeints, the ellipse parameters for our observation, $\mathbf{o}_{ep,t}$, can be calculated:
\begin{align}
    \label{equ:ellipse_center}
    \mathbf{c}_{e,t} & = \left[ \frac{c_t d_t - b_t e_t}{b_t^2 - a_t c_t}\ ,
                                \frac{a_t e_t - b_t d_t}{b_t^2 - a_t c_t}
                         \right]^\top, \\
    w_{e,t} & = \sqrt{\frac{2 (a_t e_t^2 + c_t d_t^2 + b_t^2 - 2 b_t d_t e_t - a_t c_t)}
                           {(b_t^2 - a_tc_t) \left( -\sqrt{(a_t-c_t)^2 + 4b_t^2} - (a_t+c_t) \right)}}, \\
    h_{e,t} & = \sqrt{\frac{2 (a_t e_t^2 + c_t d_t^2 + b_t^2 - 2 b_t d_t e_t - a_t c_t)}
                           {(b_t^2 - a_tc_t) \left( \sqrt{(a_t-c_t)^2 + 4b_t^2} - (a_t+c_t) \right)}}, \\
    \label{equ:ellipse_rotation}
    \theta_{e,t} & = \frac{1}{2} \tan^{-1} \left( \frac{2b_t}{a_t-c_t} \right),
\end{align}
for the ellipse center, width, height, and rotation respectively \cite{weisstein}.
Then the observation is represented as $\mathbf{o}_{ep,t} = \left[ \mathbf{c}_{e,t}\ w_{e,t}\ h_{e,t}\ \theta_{e,t} \right]^\top$.
An example of an ellipse parameter observation is shown in Fig. \ref{fig:ellipse_feature_example}. 
Note that although the coefficients of the ellipse equation, $\left[ a_t\ b_t\ c_t\ d_t\ e_t \right]^\top$, can also be an observation, $\mathbf{o}_{ep,t}$ parameterization makes it more intuitive when choosing a proper covariance of the observation noise.

\subsection{Observation Model}
To utilize the suture needle features, points and ellipse parameters, in the Bayesian filters, observation models are required.
The observation models define the relationship between the pose of the suture needle, $\mathbf{p}_t$, and the features.
In the coming subsections, we define observation models for points, points fitting to an ellipse, and ellipse parameters obtained from $\mathbf{p}_t$. 
The two point observation models and ellipse parameter observation model are used to update the Bayesian filters with detected needle points and ellipse parameters respectively.

\subsubsection{Point Observation}
The point observation model for the $i$-th feature point is calculated by projecting the $i$-th feature point's 3D location on the needle onto the image plane using the pinhole camera model.
Written explicitly, the observation model is:
\begin{equation}
    \hat{\mathbf{o}}_{p,t,i} (\mathbf{p}_t) = 
    \frac{1}{z} \mathbf{K}
    \mathbf{T}(\mathbf{p}_t) \overline{\mathbf{d}^i},
    \label{equ:ob_model_point}
\end{equation}
where $ \frac{1}{z} \mathbf{K}$ is the pinhole camera projection operator, $\mathbf{T}(\mathbf{p}_t) \in SE(3)$ is the homogeneous transform of the suture needle pose, and $\mathbf{d}^i \in \mathbb{R}^3$ is the point's 3D location on the suture needle.
Note that $\overline{\cdot}$ converts a point to its homogeneous representation, i.e. $\overline{\mathbf{d}} = [ \mathbf{d} \ 1 ]^\top$.
The observation noise is assumed zero-mean Gaussian with covariance $\Sigma_{o,p,i}\in \mathbb{R}^{2 \times 2}$:
\begin{equation}
    \label{eq:point_features_noise}
    \mathbf{o}_{p,t,i} \sim \mathcal{N}\left( \hat{\mathbf{o}}_{p,t,i} (\mathbf{p}_t), \Sigma_{o,p,i} \right),
\end{equation}
since Gaussian noise is the more frequently occurring noise in images~\cite{boncelet2009image}. 
Note that the point observation model requires the knowledge of the correspondence between the detected feature point and its 3D position, $\mathbf{d}^i$, on the suture needle.
We found that this requirement is only easy to meet for the tail and tip points of a suture needle, and hence they are what we use for our suture needle tracking.

\begin{table*}[t]
\begin{subtable}{\textwidth}
\centering
\vspace{2mm}
\begin{tabular}{c|cc|cc|cc}
    \hline
    & \multicolumn{2}{c|}{$\sigma = 0.5$} & \multicolumn{2}{c|}{$\sigma = 1$} & \multicolumn{2}{c}{$\sigma = 1.5$}\\
    \hline
    Observation & Pos error (mm) & Ori error (deg) & Pos error (mm) & Ori error (deg) & Pos error (mm) & Ori error (deg) \\
    \hline\hline
    Pose~\cite{ferro2017vision,sen2016automating} & $8.11 \pm 1.47$ & $1.26 \pm 0.21$ & $8.77 \pm 1.81$ & $1.42 \pm 0.26$ & $12.65 \pm 4.44$ & $2.06 \pm 0.58$ \\
    FPS~\cite{kurose2013preliminary} & $2.91 \pm 1.17$ & $0.55 \pm 0.1$ & $3.33 \pm 1.21$ & $0.56 \pm 0.05$ & $2.34 \pm 0.4$ & $0.52 \pm 0.07$ \\
    NCCS~\cite{ozguner2018three} & $5.03 \pm 2.65$ & $1.91 \pm 0.43$ & $11.14 \pm 5.4$ & $2.35 \pm 0.82$ & $10.56 \pm 1.84$ & $2.12 \pm 0.25$ \\
    1 point + EP & $2.86 \pm 0.61$ & $0.39 \pm 0.12$ & $3.55 \pm 0.92$ & $0.45 \pm 0.06$ & $3.58 \pm 0.34$ & $0.57 \pm 0.11$ \\
    2 points + EP & $2.41 \pm 0.35$ & $0.36 \pm 0.05$ & $2.9 \pm 0.35$ & $0.54 \pm 0.04$ & $4.14 \pm 0.65$ & $0.79 \pm 0.14$ \\
    1 point + EM & $2.19 \pm 0.24$ & $0.16 \pm 0.01$ & $2.27 \pm 0.16$ & $0.18 \pm 0.02$ & $1.74 \pm 0.42$ & $0.2 \pm 0.03$ \\
    2 points + EM & $\mathbf{0.64 \pm 0.08}$ & $\mathbf{0.07 \pm 0.01}$ & $\mathbf{0.84 \pm 0.05}$ & $\mathbf{0.12 \pm 0.01}$ & $\mathbf{1.14 \pm 0.11}$ & $\mathbf{0.17 \pm 0.02}$ \\
    \hline
\end{tabular}
\vspace{-1mm}
\label{tab:errors_static}
\end{subtable}\\
\begin{subtable}{\textwidth}
\centering
\vspace{2mm}
\begin{tabular}{c|cc|cc|cc}
    \hline
    & \multicolumn{2}{c|}{$\sigma = 0.5$} & \multicolumn{2}{c|}{$\sigma = 1$} & \multicolumn{2}{c}{$\sigma = 1.5$} \\
    \hline
    Observation & Pos error (mm) & Ori error (deg) & Pos error (mm) & Ori error (deg) & Pos error (mm) & Ori error (deg) \\
    \hline\hline
    Pose~\cite{ferro2017vision,sen2016automating} & $10.05 \pm 3.24$ & $1.67 \pm 0.46$ & $10.66 \pm 2.38$ & $1.71 \pm 0.26$ & $11.54 \pm 2.79$ & $1.91 \pm 0.33$ \\
    FPS~\cite{kurose2013preliminary} & $2.36 \pm 0.53$ & $0.41 \pm 0.06$ & $2.85 \pm 0.75$ & $0.54 \pm 0.09$ & $3.03 \pm 0.75$ & $0.56 \pm 0.13$ \\
    NCCS~\cite{ozguner2018three} & $15.85 \pm 1.37$ & $2.67 \pm 0.16$ & $16.98 \pm 1.83$ & $2.79 \pm 0.24$ & $16.47 \pm 3.42$ & $2.72 \pm 0.55$ \\
    1 point + EP & $4.03 \pm 1.53$ & $0.59 \pm 0.21$ & $4.73 \pm 0.91$ & $0.7 \pm 0.12$ & $8.62 \pm 4.94$ & $1.29 \pm 0.65$ \\
    2 points + EP & $2.53 \pm 0.41$ & $0.54 \pm 0.08$ & $2.72 \pm 0.23$ & $0.61 \pm 0.09$ & $3.1 \pm 0.47$ & $0.72 \pm 0.2$ \\
    1 point + EM & $1.38 \pm 0.25$ & $0.19 \pm 0.03$ & $1.94 \pm 0.45$ & $0.26 \pm 0.05$ & $2.34 \pm 0.68$ & $0.74 \pm 0.55$ \\
    2 points + EM & $\mathbf{0.87 \pm 0.25}$ & $\mathbf{0.12 \pm 0.03}$ & $\mathbf{0.85 \pm 0.11}$ & $\mathbf{0.13 \pm 0.01}$ & $\mathbf{1.16 \pm 0.09}$ & $\mathbf{0.2 \pm 0.01}$ \\
    \hline
\end{tabular}
\vspace{-1mm}
\label{tab:errors_moving}
\end{subtable}
\caption{Errors of tracking a static needle (upper table) and a moving needle (lower table) with 5 detected feature points. 
$\sigma$ (in pixels) is the standard deviation of the noise added to the detected needle points in the simulation environment to imitate noisy detections in real-world scenarios.
The results show that our proposed observation model, \textit{2 points + EM}, achieves the lowest tracking error in all cases.}
\label{tab:errors_simulation}
\end{table*}

\subsubsection{Points Matching to Ellipse Observation}
\label{subsubsec:ellipse_matching}

We propose a novel ellipse matching observation model for the detected points on the suture needle where their corresponding 3D locations are unknown.
The suture needle is known to project to an ellipse, and this observation model matches the detected needle points onto the suture needle's projected ellipse.
Let $\hat{a}(\mathbf{p}_t), \hat{b}(\mathbf{p}_t), \hat{c}(\mathbf{p}_t), \hat{d}(\mathbf{p}_t), \hat{e}(\mathbf{p}_t)$ be the ellipse general coefficients from (\ref{eq:general_ellipse_eq}) by projecting the pose of the circular suture needle $\mathbf{p}_t$ (see \cite{espiau1992new} for equations of projecting a circle, i.e. our suture needle, onto the image plane).
We define our matching detection as:
\begin{equation}
    \label{equ:fitted_ellipse_observation}
    \mathbf{o}_{em,t} = \hat{a}x_{t,i}^2 + 2\hat{b}x_{t,i}y_{t,i} +  \hat{c}y_{t,i}^2 + 2\hat{d}x_{t,i}+  2\hat{e}y_{t,i} \\+ 1, 
\end{equation}
by plugging in the detected needle points, $\mathbf{o}_{p,t,i} = [x_{t,i} \ y_{t,i}]^\top$, into the ellipse equation. Note that $(\mathbf{p}_t)$ is dropped from the ellipse coefficients for concise notation.
The noise on this matching error can be derived using (\ref{eq:point_features_noise}) as $\mathbf{o}_{em,t}$ is a derived Random Variable from $\mathbf{o}_{p,t,i}$.
We approximate $\mathbf{o}_{em,t}$ as Gaussian with variance $\sigma^2_{em,i}$:
\begin{equation}
    \label{equ:ob_model_deviation}
    \mathbf{o}_{em,t,i} \sim \mathcal{N}\left( 0,  \sigma^2_{em,i} \right),
\end{equation}
since Gaussian noise is the more frequently occurring noise in images~\cite{boncelet2009image}.
The mean is 0 because taking the expectation of (\ref{equ:fitted_ellipse_observation}) yields plugging $\hat{\mathbf{o}}_{p,t,i}$ into the general ellipse equation, (\ref{eq:general_ellipse_eq}), with coefficients $\hat{a}, \hat{b}, \hat{c}, \hat{d}, \hat{e}$. 
This is guaranteed to be 0 since $\hat{\mathbf{o}}_{p,t,i}$ is defined to always be on the suture needle and hence on the ellipse.


The variance of the observation noise for matching each detected point to the projected ellipse, $\sigma_{em,i}^2$, can be derived by the noise modeled on the detected point features, $\mathbf{o}_{p,t,i}$, in (\ref{eq:point_features_noise}) with a few additional approximations.
To show this, we first consider how the noise in the pixel frame results in the noise of the ellipse matching observation. 
With the estimated ellipse coefficients, $[ \hat{a}_t\ \hat{b}_t\ \hat{c}_t\ \hat{d}_t\ \hat{e}_t ]^\top$, and the noisy detected feature point, $\left[ x_{t,i}\ y_{t,i} \right]^\top$, satisfying the following equations: 
\begin{align}
    & x_{t,i} = \hat{x}_{t,i} + v_x, \quad y_{t,i} = \hat{y}_{t,i} + v_y, \\
    & \hat{a}_t \hat{x}_{t,i}^2 + 2 \hat{b}_t \hat{x}_{t,i} \hat{y}_{t,i} + \hat{c}_t \hat{y}_{t,i}^2 + 2 \hat{d}_t \hat{x}_{t,i} + 2 \hat{e}_t \hat{y}_{t,i} + 1 = 0, \\
    & v_x \sim \mathcal{N}\left(0, \bar{\sigma}_{p,i}^2\right), \quad v_y \sim \mathcal{N}\left(0, \bar{\sigma}_{p,i}^2\right),
\end{align}
the ellipse matching observation becomes 
\begin{align}
    & \hat{a}_t x_{t,i}^2 + 2 \hat{b}_t x_{t,i} y_{t,i} + \hat{c}_t y_{t,i}^2 + 2 \hat{d}_t x_{t,i} + 2 \hat{e}_t y_{t,i} + 1 \notag\\
    =\ & \hat{a}_t \hat{x}_{t,i}^2 + 2 \hat{b}_t \hat{x}_{t,i} \hat{y}_{t,i} + \hat{c}_t \hat{y}_{t,i}^2 + 2 \hat{d}_t \hat{x}_{t,i} + 2 \hat{e}_t \hat{y}_{t,i} + 1 \notag\\
    & + \hat{a}_t v_x^2 + 2 \hat{b}_t v_x v_y + \hat{c}_t v_y^2 
      + 2 \left( \hat{a}_t \hat{x}_{t,i} + \hat{b}_t \hat{y}_{t,i} + \hat{d}_t \right) v_x \notag\\
    & + 2 \left( \hat{b}_t \hat{x}_{t,i} + \hat{c}_t \hat{y}_{t,i} + \hat{e}_t \right) v_y \notag\\
    =\ & \hat{a}_t v_x^2 + 2 \hat{b}_t v_x v_y + \hat{c}_t v_y^2 
      + 2 \left( \hat{a}_t \hat{x}_{t,i} + \hat{b}_t \hat{y}_{t,i} + \hat{d}_t \right) v_x \notag\\
    & + 2 \left( \hat{b}_t \hat{x}_{t,i} + \hat{c}_t \hat{y}_{t,i} + \hat{e}_t \right) v_y \notag\\
    =\ & d\left( v_x, v_y \right), 
    \label{equ:func_ob_deviation}
\end{align}
where $\left[ \hat{x}_{t,i}\ \hat{y}_{t,i} \right]^\top \in \mathbb{R}^2$ is the expected $i$-th feature point, and $v_x \in \mathbb{R}$ / $v_y \in \mathbb{R}$ is the noise of the x / y coordinate in the pixel frame from the $i$-th detected feature point. 
We also assume $v_x$ and $v_y$ are independent. 

By definition, the variance of equation (\ref{equ:func_ob_deviation}) is 
\begin{equation}
    \mathbb{V} \left[ d\left( v_x, v_y \right) \right] = 
    \mathbb{E} \left[ \left( d\left( v_x, v_y \right) 
               - \mathbb{E} \left[ d\left( v_x, v_y \right) \right] \right)^2 \right].
\end{equation}
The function $d\left( v_x, v_y \right)$ can be approximated by the first-order Taylor expansion: 
\begin{align}
    \mathbb{E} \left[ d\left( v_x, v_y \right) \right] 
    & \approx \mathbb{E} \left[ d(0,0) + d_x(0,0) v_x + d_y(0,0) v_y \right] \notag\\
    & = d(0,0), 
\end{align}
and 
\begin{align}
    & \left( d\left( v_x, v_y \right) - d(0,0) \right)^2 \notag\\
    & \approx \left( d(0,0) + d_x(0,0) v_x + d_y(0,0) v_y - d(0,0) \right)^2 \notag\\
    & = d_x(0,0)^2 v_x^2 + 2 d_x(0,0) d_y(0,0) v_x v_y + d_y(0,0)^2 v_y^2, 
\end{align}
where $d_x(0,0) = \frac{\partial d}{\partial v_x}(0,0)$, $d_y(0,0) = \frac{\partial d}{\partial v_y}(0,0)$. 
Thus, 
\begin{align}
    & \mathbb{V} \left[ d(v_x, v_y) \right] \notag\\
    \approx\ & \mathbb{E} \left[ d_x(0,0)^2 v_x^2 + 2 d_x(0,0) d_y(0,0) v_x v_y + d_y(0,0)^2 v_y^2 \right] \notag\\
    =\ & d_x(0,0)^2 \mathbb{E} \left[ v_x^2 \right] 
         + 2 d_x(0,0) d_y(0,0) \mathbb{E}\left[ v_x, v_y \right] \notag\\
    &    + d_y(0,0)^2 \mathbb{E} \left[ v_y^2 \right] \notag\\
    =\ & d_x(0,0)^2 \mathbb{V} \left[ v_x \right] 
         + 2 d_x(0,0) d_y(0,0) \mathbb{C}\left[ v_x, v_y \right] \notag\\
    &    + d_y(0,0)^2 \mathbb{V} \left[ v_y \right] \notag\\
    =\ & d_x(0,0)^2 \bar{\sigma}_{p,i}^2 + d_y(0,0)^2 \bar{\sigma}_{p,i}^2 
    \label{equ:cov_equals_0}\\
    =\ & 4 \left[ \left( \hat{a}_t \hat{x}_{t,i} + \hat{b}_t \hat{y}_{t,i} + \hat{d}_t \right)^2 
                  + \left( \hat{b}_t \hat{x}_{t,i} + \hat{c}_t \hat{y}_{t,i} + \hat{e}_t \right)^2 \right] \bar{\sigma}_{p,i}^2, 
    \label{equ:derived_variance}
\end{align}
where $\mathbb{C}(v_x, v_y)$ is the covariance of $v_x$ and $v_y$. 
The covariance $\mathbb{C}(v_x, v_y) = 0$ since $v_x$ and $v_y$ are independent.
Finally, we approximate $\hat{x}_{t,i}, \hat{y}_{t,i} \approx x_{t,i}, y_{t,i}$ to complete the computation.
Therefore, (\ref{equ:derived_variance}) shows how the variance of the observation noise $\sigma_{em,i}^2$ can be represented by the variance in the pixel frame, $\bar{\sigma}_{p,i}^2$. 

Note that this model can work when the needle is partially occluded since it allows only one detected point as the input for fitting to a projected ellipse. 



\subsubsection{Ellipse Parameter Observation}

The ellipse parameter observation model is defined by projecting the circular suture needle to an ellipse, $\hat{a}(\mathbf{p}_t), \hat{b}(\mathbf{p}_t), \hat{c}(\mathbf{p}_t), \hat{d}(\mathbf{p}_t), \hat{e}(\mathbf{p}_t)$ (see \cite{espiau1992new} for equations of projecting a circle, i.e. our suture needle, onto the image plane).
These coefficients are converted to the ellipse parameterization using (\ref{equ:ellipse_coeff_fitting})-(\ref{equ:ellipse_rotation}), giving the observation model $\hat{\mathbf{o}}_{ep,t} (\mathbf{p}_t) = [ \hat{\mathbf{c}}_{e,t}(\mathbf{p}_t)\ \hat{w}_{e,t}(\mathbf{p}_t)\ \hat{h}_{e,t}(\mathbf{p}_t)\ \hat{\theta}_{e,t}(\mathbf{p}_t) ]^\top$.
The observation model noise is assumed Gaussian with covariance $\Sigma_{ep,t}$:
\begin{equation}
    \mathbf{o}_{ep,t} \sim \mathcal{N}\left( \hat{\mathbf{o}}_{ep,t}, \Sigma_{ep,t} \right),
\end{equation}
since Gaussian noise generalizes well over a wide range of distributions.

\section{Experiment and Results}

\subsection{Simulated Tracking Experiments}

The proposed observation models are first evaluated in a CoppeliaSim~\footnote{https://www.coppeliarobotics.com/} simulation environment used in our previous work~\cite{chiu2020bimanual}. 
In this environment, a radius 5.4mm suture needle is initially held in a robotic gripper. 
Also, a stereo camera model is added to get the observation features from images with a size of $256 \times 256$. 
For each image that contains a suture needle, five feature points are detected and distributed as shown in Fig. \ref{fig:point_feature_example}. 
To imitate the noisy detected feature points in real-world scenarios, all projected points in the simulation environment are added with a noise sampled from $\mathcal{N}(\mathbf{0}, \Sigma)$, where $\Sigma \in \mathbb{R}^{2 \times 2}$ is a diagonal matrix with each diagonal element being $\sigma^2$. 

The observation features compared include 
\begin{enumerate}
    \item \textit{Pose~\cite{ferro2017vision, sen2016automating}}: the pose of the needle reconstructed from the fitted ellipse and the detected tail point, 
    \item \textit{Feature point similarity (FPS)~\cite{kurose2013preliminary}}: all the detected needle points (with their predefined 3D-point registration), 
    \item \textit{Normalized cross-correlation similarity (NCCS)~\cite{ozguner2018three}}: the normalized cross-correlation similarity between the detected and the virtually rendered needle images is calculated and used to update the weight of each particle, 
    \item \textit{1 point + ellipse parameters (EP)}: the detected tail point + the fitted ellipse parameters, 
    \item \textit{2 points + EP}: the detected tail and tip points + the fitted ellipse parameters, 
    \item \textit{1 point + ellipse matching (EM)}: the detected tail point + the ellipse matching observation (Section \ref{subsubsec:ellipse_matching}) for 4 other points, and 
    \item \textit{2 points + EM}: the detected tail and tip points + the ellipse matching observation for 3 other points. 
\end{enumerate}
Note that both the ellipse-parameter and ellipse-matching observations require at least one point observation to anchor the orientation of the needle since the projected ellipse stays identical if the needle only rotates along the axis perpendicular to the image. 
These observation features are used in PF with 5000 particles to track the needle with two different motions: static and moving along a predefined trajectory.
The initialization of the needle pose before tracking is obtained from the pose reconstruction method proposed in~\cite{lo2002trip}, so the needle tracking process can be done fully automatically without any knowledge of the initial pose. 

Table \ref{tab:errors_simulation} shows the pose errors of tracking a suture needle with only five detected needle points. 
The results from all tables suggest that the \textit{Pose} and \textit{NCCS} observations lead to larger tracking errors. 
This is because with only five detected points: 
(i) the fitted ellipse can be wildly inaccurate, which primarily affects the accuracy of the reconstructed pose; 
(ii) the needle images rendered are too unrealistic to be compared fairly with another image, especially when the registration between the detected points and their 3D correspondence is inaccurate. 

The \textit{FPS} and \textit{Point + EP} observations lower the tracking errors of position to be smaller than 5mm and that of orientation to be smaller than 1 degree. 
Nevertheless, with imperfect 3D positions registered from the detected points, especially from the points other than the tail or tip, the tracking errors of the \textit{FPS} observation cannot be adequately reduced. 
On the other hand, since the inaccurately fitted ellipse appears very frequently during tracking, the average tracking error is difficult to decrease further. 

The \textit{Point + EM} observation shows great improvement compared to the \textit{Pose}, \textit{FPS}, \textit{NCCS}, and \textit{Point + EP} observations. 
It achieves the lowest average tracking error for both position and orientation in all cases. 
An example of the projection of the tracked pose is shown in Fig. \ref{fig:ellipse_tracked_pose}.
Moreover, the average computational time per frame for this observation is around 0.03s, faster than \textit{NCCS} (which takes several seconds) and comparable to other observations (which take about 0.02s). 
This suggests that the \textit{Point + EM} observation is more robust to the noise in the detections and does not suffer from the trade-off between accuracy and computational time. 

\subsection{Automated Suture Needle Passing}
We also evaluated the proposed needle tracking method with the needle regrasping policy developed in our previous work~\cite{chiu2020bimanual} in the simulation environment. 
The noise added to the detected feature points is sampled from $\mathcal{N}(\mathbf{0}, \Sigma)$, where $\Sigma \in \mathbb{R}^{2 \times 2}$ is a diagonal matrix with each diagonal element being $\sigma^2$.
The pose of the needle is tracked with the \textit{2 points + EM} observation. 
Each experiment is run for 30 trials. 

\begin{table}[t]
\centering
\vspace{2mm}
\begin{tabular}{c|cccccc}
    \hline
    $\sigma$ & $0$ & $0.4$ & $0.8$ & $1.2$ & $1.6$ & $2$ \\
    \hline\hline
    Success rate ($\%$) & $100$ & $100$ & $93.33$ & $96.67$ & $86.67$ & $80$ \\
    \hline
\end{tabular}
\vspace{-0.5mm}
\caption{Success rate of running the needle regrasping policy~\cite{chiu2020bimanual} when using our proposed method to localize the needle. 
$\sigma$ (in pixels) is the standard deviation of the noise added to the detected needle points in the simulation environment. 
The success rate remains high in a noisy environment, indicating our needle tracking method can effectively localize the needle while being manipulated.}
\label{tab:regrasping_results}
\end{table}

Table \ref{tab:regrasping_results} shows the success rate of running the needle regrasping policy with tracked needle poses. 
With the proposed observation model, the success rate of needle regrasping under uncertainty remains high (over $80\%$) even when the environmental noise parameter $\sigma$ increases to $2$. 
This demonstrates that the proposed method can effectively track the uncertainty in the observations. 

\subsection{Real World Tracking Experiment}
To evaluate our proposed methods in real-world scenarios, the \textit{2 points + EM} observation for PF with 2000 particles is tested on suture needles with radii $7$mm and $11.5$mm. 
During tracking, a needle is grasped by a Patient Side Manipulator (PSM) arm from the da Vinci Research Kit (dVRK)~\cite{kazanzides2014open} using a Large Needle Driver (LND). 
The images for feature extraction are captured by dVRK's stereo-endoscopic camera, which is 1080p and runs with 30fps. 
The actions are provided by tracking the kinematic changes of the PSM arm using our previous method~\cite{richter2021robotic}. 

We use DeepLabCut (DLC)~\cite{mathis2018deeplabcut}, the state-of-the-art keypoint detector, to obtain markerless detections of the needle. 
Previous work has demonstrated that DLC achieves accurate detections with less than 50 manually labeled samples~\cite{lu2020super}, so minimal training is required to adapt DLC to our new scenario of suture needles. 

\begin{figure}[t]
    \centering
    \vspace{2mm}
    \includegraphics[width=0.48\textwidth]{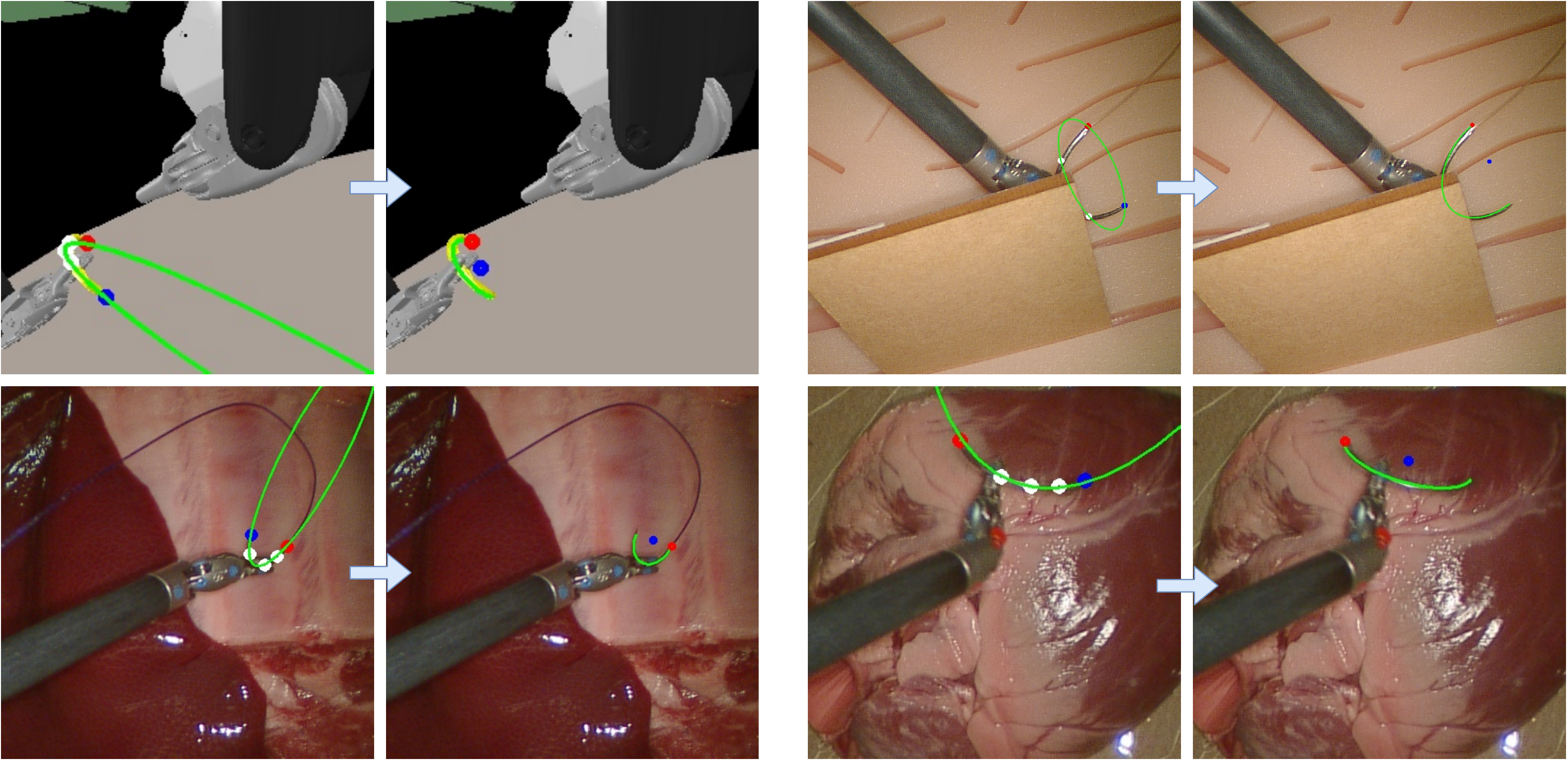}
    \caption{The detected feature points, inaccurately fitted ellipse colored in green (arrow's left), and the projection of the tracked pose (arrow's right).
    The high consistency between the tracked projection and the real one indicates that our tracking method is not affected by the inaccuracy of the fitted ellipse, unlike previous methods that rely on accurately fitting an ellipse to the detected points~\cite{iyer2013single,ferro2017vision,sen2016automating}.}
    \vspace{-0.5mm}
    \label{fig:ellipse_tracked_pose}
\end{figure}

The tracked needle pose is projected onto the image to demonstrate that the proposed method effectively tracks a suture needle. 
Then, the overlap between the tracked pose and the actual pose projections indicates if the result is accurate enough. 
Fig. \ref{fig:cover_image} and \ref{fig:ellipse_tracked_pose} show the projection of the tracked pose under different scenarios, including when a needle is in ex vivo environments or under occlusion. 
It can be observed that the tracked pose overlaps well with the actual pose in most of the cases, even when the fitted ellipse is wildly inaccurate (Fig. \ref{fig:ellipse_tracked_pose}). 
Unlike the reconstructed pose and ellipse parameter observations, which primarily rely on the accuracy of the detected ellipse parameters, our proposed observation model is not affected by their accuracy.

\section{Discussion and Conclusion}

In this work, we present a robust tracking method for estimating the 6D pose of a suture needle.
We present multiple observation models based on endoscopic detections of a suture needle.
The observation models incorporate uncertainty that comes with markerless needle detections. 
Our results show that using the \textit{Point and Points Matching to Ellipse} observation models (\textit{Points + EM}) outperforms previous approaches in suture needle localization and can work under occlusion and ex vivo environments.
We credit the \textit{Points + EM} observation model's performance because its uncertainty is derived from a pixel-level distribution, and it requires minimal association with 3D points on the needle.
For future work, our proposed method will be combined with constrained Bayesian estimation~\cite{shao2010constrained} to constrain the suture needle's pose when grasped by a surgical tool.


\balance
\bibliographystyle{IEEEtran}
\bibliography{ref}

\end{document}